\documentclass{article}

\PassOptionsToPackage{numbers, compress}{natbib}

\usepackage[preprint]{sty}

\usepackage[utf8]{inputenc} 
\usepackage[T1]{fontenc}    
\usepackage{hyperref}       
\usepackage{url}            
\usepackage{booktabs}       
\usepackage{amsfonts}       
\usepackage{nicefrac}       
\usepackage{microtype}      
\usepackage{xcolor}         
\usepackage{graphicx}
\usepackage{amsmath}
\usepackage{array}
\usepackage{diagbox}
\usepackage{multirow}

\usepackage{algorithm}
\usepackage{algpseudocode}
\usepackage{amsmath}

\usepackage{float}
\usepackage{subfig}

\usepackage{xcolor}

\newcolumntype{L}[1]{>{\raggedright\arraybackslash}m{#1}}
\newcolumntype{C}[1]{>{\centering\arraybackslash}m{#1}}
\newcolumntype{R}[1]{>{\raggedleft\arraybackslash}m{#1}}

\title{PTA: Enhancing Multimodal Sentiment Analysis through Pipelined Prediction and Translation-based Alignment}

\author{
        Shezheng Song\textsuperscript{\rm 1}, 
        Shasha Li\textsuperscript{\rm 1}, 
        Shan Zhao\textsuperscript{\rm 2},
        Chengyu Wang\textsuperscript{\rm 1},
        Xiaopeng Li\textsuperscript{\rm 1},
        Jie Yu\textsuperscript{\rm 1},\\  
        \bf
        Qian Wan\textsuperscript{\rm 4} 
        Jun Ma\textsuperscript{\rm 1}, 
        Tianwei Yan\textsuperscript{\rm 1},
        Wentao Ma\textsuperscript{\rm 3},
        Xiaoguang Mao\textsuperscript{\rm 1}
        \\
        betterszsong@gmail.com \\
        \textsuperscript{\rm 1}National University of Defense Technology\\
        \textsuperscript{\rm 2}Hefei University of Technology \\
        \textsuperscript{\rm 3}Anhui Agricultural University \\
        \textsuperscript{\rm 4}Central China Normal University
}    

\begin{document}

\maketitle

\begin{abstract}

    Multimodal aspect-based sentiment analysis (MABSA) aims to understand opinions in a granular manner, advancing human-computer interaction and other fields. Traditionally, MABSA methods use a joint prediction approach to identify aspects and sentiments simultaneously. However, we argue that joint models are not always superior. Our analysis shows that joint models struggle to align relevant text tokens with image patches, leading to misalignment and ineffective image utilization.
    In contrast, a pipeline framework first identifies aspects through MATE (Multimodal Aspect Term Extraction) and then aligns these aspects with image patches for sentiment classification (MASC: Multimodal Aspect-Oriented Sentiment Classification). This method is better suited for multimodal scenarios where effective image use is crucial. We present three key observations: (a) MATE and MASC have different feature requirements, with MATE focusing on token-level features and MASC on sequence-level features; (b) the aspect identified by MATE is crucial for effective image utilization; and (c) images play a trivial role in previous MABSA methods due to high noise.
    Based on these observations, we propose a pipeline framework that first predicts the aspect and then uses translation-based alignment (TBA) to enhance multimodal semantic consistency for better image utilization. Our method achieves state-of-the-art (SOTA) performance on widely used MABSA datasets Twitter-15 and Twitter-17. This demonstrates the effectiveness of the pipeline approach and its potential to provide valuable insights for future MABSA research.
    For reproducibility, the code and checkpoint will be released.

\end{abstract}

\section{Introduction}
\label{sec:intro}

\begin{figure}[htbp]
            \centering
            \subfloat[Joint prediction]{\includegraphics[height=.22\linewidth]{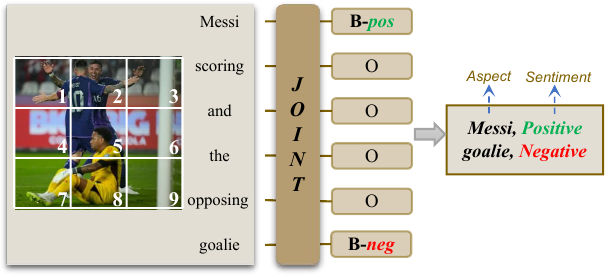}\label{fig:intro1}}
            \hspace{2mm}
            \subfloat[Pipeline framework]{\includegraphics[height=.22\linewidth]{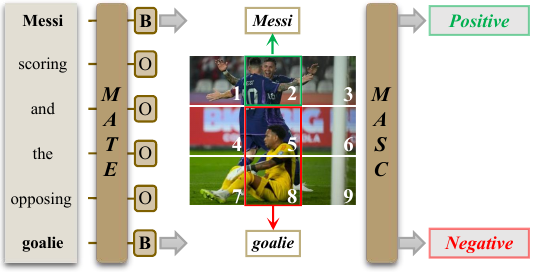}\label{fig:intro2}}
            \caption{Frameworks of MABSA. "B-pos" means "Begin, Positive"; I and O represent "In" and "Out" respectively, indicating the aspect position.
            In pipeline framework, MATE first identifies the aspect as "Messi." Then, "Messi" interacts with the image for sentiment classification "Positive."
            }
        \end{figure}

    Multimodal aspect-based sentiment analysis (MABSA) refers to the process of understanding opinions in human expression in a more granular manner. Research in sentiment analysis facilitates the development of human-computer interaction~\citep{humanInteraction}, healthcare~\citep{health} and etc.
    As shown in Figure \ref{fig:intro1}, MABSA is typically achieved through joint prediction of aspects and their corresponding sentiments, such as "Messi:B-pos". In fact, MABSA could also be decomposed into a pipeline process, which comprises two subtasks: MATE (multimodal aspect term extraction) and MASC (multimodal aspect-oriented sentiment classification). As depicted in Figure \ref{fig:intro2}, after extracting aspect "Messi" from text~\cite{wang2021}(MATE), MASC could assign sentiment label "positive" to the given aspect "Messi"~\cite{Ho2022, Zhang_Wang_Zhang_2021}. 
 
    The joint prediction mode is widely adopted by existing MABSA methods~\cite{UMT, OSCGA, JML, VLP, CMMT, DTCA, m2df, DQPSA}. While there is a long-held belief that \textit{joint models can better capture the semantics of aspect and sentiment}, we argue that this assumption does not universally hold. In these joint models, each token is annotated as one of [B-positive, B-negative, B-neutral, I, O], thereby jointly annotating the sentence with (aspect, sentiment) tuples. Although this approach can reduce error accumulation, it poses significant challenges in determining the relevant text tokens (aspect) before completing the joint prediction.
    Due to the uncertainty regarding which tokens in the text are useful, as illustrated in Figure \ref{fig:intro1}, selecting image patch 2 and aligning it with the useful token for predicting "B-pos" is difficult. Consequently, the difficulty in identifying the image patches related to the aspect leads to misalignment of multimodal information and ineffective image utilization.
    In contrast, the pipeline framework can first identify the aspect (MATE), and then align the aspect with the corresponding image patches, enabling more precise and accurate image utilization. The pipeline is better suited for multimodal scenarios where the effective use of images is crucial.

    Through the analysis of the current MABSA methods, we observe that 
    \textbf{(a)} 
    \textbf{The MATE and MASC models have different requirements for features}. 
    As shown in Figure \ref{fig:intro2}, MATE focuses on the label results for each token, emphasizing \textit{textual token-level features}, such as "Messi". MASC, on the other hand, focuses on classification results of \textit{local sentences and images}, emphasizing sequence-level features, such as "Messi scoring" and image patch 2. Sharing MATE and MASC can lead to confusion in feature representation for joint models, damaging performance.
    \textbf{(b)} 
    \textbf{The aspect of MATE results is crucial for utilizing images}. 
    In the joint prediction approach of MABSA, techniques like attention are used to map patches in images to tokens in text. However, in cases where it’s uncertain which token in the text is relevant, this mapping approach severely limits the utilization of images. 
    In conclusion, we argue that pipeline framework could capture features at different levels, including token-level and sentence-level. Moreover, it can utilize predicted aspects to capture effective image information more finely. Therefore, we believe that pipeline approach is better than joint framework.
    \textbf{(c)}
    \textbf{Images play a trivial role in previous MABSA methods}~\cite{DTCA, DQPSA, m2df, VLP, AoM}.
    In fact, there is a significant degree of randomness in how people post opinions. In the process, sentiment expression primarily relies on text, while accompanying images are often chosen randomly.
    Due to the data for sentiment analysis tasks often coming from social media platforms~\citep{TwitterLabel, TwitterDataset, TwitterPropose}, the image data for MABSA contains more noise compared with other tasks~\citep{maitr, dwe, wang2022progressive}.
    \textit{Even if the aspect is determined, the high noise in the image makes it difficult to map the relevant image patches with the aspect}. High-noise image information requires a higher demand for how models utilize images. Directly using images may even lead to performance degradation. Simply using attention mechanisms~\cite{DTCA, CMMT} or mapping features to latent space~\citep{DQPSA, JML, AoM, VLP} for alignment is not advisable because the noise in images may prevent effective information expression. 
    Finally, we conduct experimental validation to support our observations and corresponding analysis (\S \ref{sec:observation-experiment}).

    Based on observations \textbf{(a)} and \textbf{(b)}, we modify the framework of the MABSA task, replacing the joint prediction framework with a pipeline framework where the aspect is predicted first (MATE), followed by combining relevant parts from the image to complete sentiment classification (MASC). The MATE is considered a token-level label task, predicting the BIO labels for each token and obtaining aspects based on the prediction results. Then, in MASC, the aspect obtained in the MATE is used to extract relevant visual features;
    Based on observation \textbf{(c)}, we aim to strengthen the consistent parts between the text and images from a semantic coherence perspective. Specifically, we leverage semantic consistency among modalities, much like in translation where the inherent semantics conveyed by different languages are consistent. Different modalities are treated as different languages, transforming the multimodal alignment issue into a multilingual translation problem. 
    In detail, vision-to-text(\textit{v2t}) translation is leveraged to learn the visual semantic consistency with text.
    Through the translation-based alignment (TBA), we direct the model's attention towards the consistent visual information with text, which often represents the core of what multimodal information aims to convey.

    The contributions are summarized as follows:
    \begin{itemize}
        \item We highlight three experimentally-supported observations and the corresponding insights in MABSA, which hold the potential to offer valuable insights for subsequent research and for tasks employing joint prediction, e.g., Multimodal Named Entity Recognition (MNER).
        \item  We replace the widely used joint prediction framework with a pipeline framework for MABSA, where MATE is completed before MASC, providing a fresh perspective for solving MABSA and validating its effectiveness. 
        \item  In MASC task of pipeline framework, drawing inspiration from machine translation, we consider different modalities as different languages and propose a translation-based alignment (TBA) method to address the issue of multimodal information alignment. 
        \item Our method achieves State-of-The-Art (SOTA) performance in public and widely adopted MABSA datasets: Twitter-15 and Twitter-17. Our method also outperforms Large Language Model (LLM) Llama2, ChatGPT, and VisualGLM6B.
    \end{itemize}

\section{Related Work}

    Most previous MABSA methods~\citep{UMT, OSCGA, JML, VLP, CMMT, DTCA, m2df, DQPSA} rely on joint prediction, labeling text tokens while simultaneously predicting aspects and sentiments. For instance, DQPSA~\citep{DQPSA} employs a prompt and an energy-based pairwise expert module to improve the prediction of joint prediction based on pairwise stability. JML~\citep{JML} jointly handles multimodal aspect extraction and sentiment analysis, incorporating cross-modal relation detection. DTCA~\cite{DTCA} uses tasks to bridge modality gaps and an adaptive multimodal mechanism to dynamically adjust modality importance.
    Additionally, a few methods like AoM~\citep{AoM} recognize the importance of aspects for the MASC task and use Spacy to label all nouns as candidate aspects. However, part-of-speech tagging is too coarse, and the extracted candidate aspects can only serve as supplementary information.

    Joint sentiment analysis methods struggle to identify useful text. This limitation is not as pronounced in text tasks but becomes significant in multimodal scenarios, where effective text is needed to help determine relevant images. Traditional joint methods interact with images using coarse-grained sentence-level information, leading to suboptimal results:
    DTCA~\cite{DTCA} attempts to use image information through two auxiliary tasks: vision-aware extraction and token-patch matching. M2DF~\cite{m2df} aims to mitigate the negative impact of noisy images by proposing a noise reduction algorithm. VLP~\cite{VLP} introduces additional image-related pretraining tasks to enhance image utilization, with high training costs. CMMT~\cite{CMMT} extracts adjective-noun pairs (ANPs)~\citep{ANPsutilization, ANPsextractor} from images as additional information. 
    Despite these various approaches, ablation studies reveal that they do not effectively utilize images~\cite{DTCA, m2df, VLP, CMMT, DQPSA}. Therefore, an effective method for utilizing images in MABSA is needed.
    In the direction of image utilization in multimodal scenarios, attention has been focused on the inherent multimodal semantic consistency~\cite {yin2019semantics}. Inspired by interlingual translations~\cite {jensen2015optimising}, generative approachs~\cite{CycleGAN, XMCGAN, chen2023learning} are taken to address the transformation and alignment between modalities.

\section{Pipeline Framework}
\label{pipeline}
    We solve the MABSA in two steps (aspect extraction \S \ref{sec:ae} and sentiment classification \S \ref{sec:sc}) and use translated-based alignment method (\S \ref{sec:TBAM}) to enhance image utilization in high-noise scenarios during sentiment classification.

    This input contains two parts of information: image and text, so it requires different processing and encoding for both types of information.  
    
    \textbf{Textual Representation.} The input text is tokenized by the WordPiece \citep{WordPiece} tokenizer as same as in the DeBERTa~\citep{DeBERTa} model to obtain token embeddings $t^0$ with a word embedding matrix $T \in \mathbb{R}^{N \times|\hat{V}|}$. 
        \begin{equation}
            t^0 = [t_{[cls]}T, t_1T, t_2T, ..., t_NT, t_{[sep]}T] \quad \in \mathbb{R}^{N\times d}
        \end{equation} 
    where $|\hat{V}|$ is the number of the vocabulary items and $N$ is the length of text. Here, the \textit{[cls]} and \textit{[sep]} tokens respectively respond to $<s>$ and $</s>$ tokens in the DeBERTa model.  
    
    \textbf{Visual Representation.} Following the ViT~\citep{VIT}, image is first sliced into a sequence of patches $v = [v_1, v_2, . . , v_M ] \in \mathbb{R}^{M \times p^2}$, where $(p, p)$ is the resolution of each patch and $M$ is the resulting number of patches. Each patch is then flattened and prepended with a special token, i.e., $v_{[cls]}$, followed by linear projection $V \in \mathbb{R} ^{ P^2\times d}$. Thus we could get the result patch embeddings $v^0 $.
        \begin{equation}
            v^0 = [v_{[cls]}, v_1V,v_2V..., v_MV]  \quad \in \mathbb{R}^{M\times d}
        \end{equation}

    \begin{figure*}[tbp]
        \centering
        \includegraphics[width=.8\textwidth]{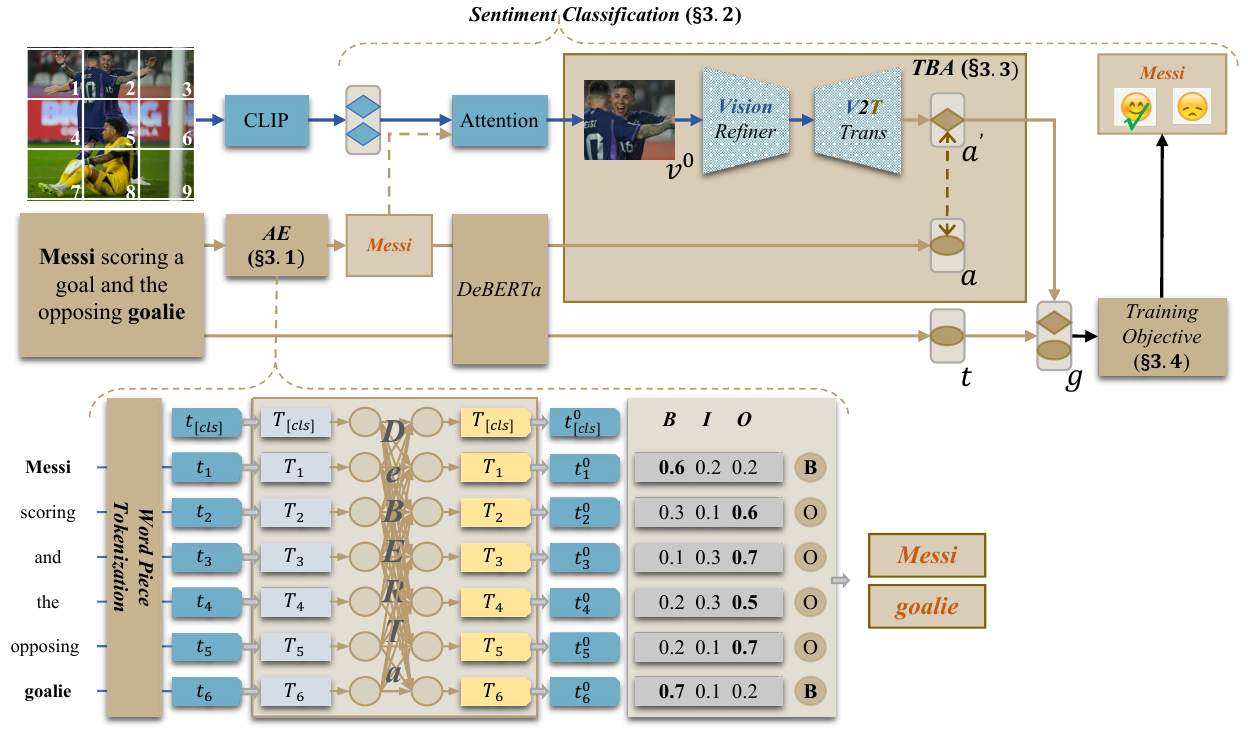}
        \caption{Overview of our method. AE means aspect extraction module, which is used to extract "Messi" from the sentence for the subsequent sentiment classification task. $a$ and $t$ represent the aspect and sentence feature, respectively. $a'$ is the translated image feature, aimed at maintaining semantic consistency with $a$. $B$ denotes the beginning of the aspect position, while $O$ indicates that the token is not an aspect.
        }
        \label{fig:model}
    \end{figure*}
    
    \subsection{Aspect Extraction}
    \label{sec:ae}

        The textual representation obtained from DeBERTa, i.e., $t^0 = [t_{[cls]}T, t_1T, t_2T, ..., t_NT, t_{[sep]}T]$ is fed to a fully connected layer with softmax activation to predict the labels for the tokens:
            \begin{equation}
                y_n = softmax(Wt_n+b) \quad \in \mathbb{R}^{N\times k}
            \end{equation}
        where $W \in \mathbb{R}^{d\times k} $ and $b \in \mathbb{R}^{N\times k}$ are trainable parameters, and $k$ is the number of the candidate tags.   
        For token label task, instead of using 7 tags ( B-pos, B-neg, B-neu, I-pos, I-neg, I-neu, O) as in previous works~\cite{DTCA}, we are only concerned with predicting the position of aspects, and thus the number of tags $k=3$ (B, I, O).
        The number of classes is compressed by half, decreasing the prediction error caused by sentiment analysis.
        
        \textbf{Aspect Label Loss.}
        The aspect label loss $\mathcal{L}_a$ is categorical cross-entropy, where $\bar{y_i}$ is the ground-truth token label (B, I, O).
            \begin{equation}
                \mathcal{L}_a   = CE(\bar{y_n}, y_n) = -\sum_{i=1}^{n} \bar{y_i} log(y_i)
            \end{equation}

        As shown in Algorithm \ref{alg:logits}, we convert the aspect logits (probabilities of B, I, O) from sequence labeling into textual aspect $ta$, such as "Messi". Then $ta$ is encoded through DeBERTa to get aspect feature $a \in \mathbb{R}^{N\times d}$.
        B, I, and O respectively stand for Begin, Inside, and Out, representing the beginning position, inside position, and outside of aspect.

    \subsection{Sentiment Classification}
    \label{sec:sc}

    As shown in Figure \ref{fig:model}, we feed the aspect information $a = AE(text)$ obtained from the previous step as additional supplementary information into the sentiment classification task of the sentence. Specifically, we use aspect information to extract relevant parts of information from the image, such as using "Messi" to apply attention mechanism to visual features related to characters in the image, to obtain $v^0 = att(a, v)$.
    We use a translation-based alignment module (\S \ref{sec:TBAM}) to transform the visual and textual information, expecting the translated information to be consistent with the target information. For example, the translated textual information $a'$ should maintain semantic consistency with the original information $a^0$. In this translation process, we strengthen the parts of the text and image information that are highly consistent and weaken the inconsistent parts.
    \begin{equation}
        a' =  \mathbb{T}_{v2t}(v^0) \quad \in \mathbb{R}^{N\times d}
    \end{equation}
    We add the translated feature $a'$ from the translation module to the aspect features $a$ (\S\ref{sec:ae}).
    The fused features $g$ are passed through linear layer for classification, thus we could get sentiment label $s$.
    \begin{align}
        g &= a'_{[cls]} + t_{[cls]} \quad \in \mathbb{R}^{1\times d} \\ 
        s &= softmax(W^gg+b^g) \quad \in \mathbb{R}^{1}
    \end{align}
    The final output $(a_i,s_i)$ consists of pairs of (1) the aspect $a$ from the MATE and (2) the corresponding sentiment $s$ obtained from the sentence-level sentiment classification task.

    \subsection{Translation-Based Alignment}
    \label{sec:TBAM}
        The TBA part in Figure \ref{fig:model} shows the proposed multimodal alignment method, named translation-based alignment method. Specifically, we incorporate a feature refiner to refine valuable information comprehensively. Then, to align multimodal features, we take a cross-modal translation-based alignment (\textbf{TBA}) method to complete the vision-to-text (\textbf{v2t}) translation.
            
        We feed the original visual feature $v^0$ into v2t translation module to get translated textual feature $a'$.
        \begin{equation}
            a' = \mathbb{T}_{v2t}(v^0)
        \end{equation}
        CLIPEncoderLayer \citep{CLIP} is taken as vision refiner $\mathcal{R}_v$ to get refined vision-related feature $\theta_v$. 
        \begin{align}
            A = softmax &( \frac{(W^Q v^0)^T (W^K v^0)}{\sqrt{d}}) (W^V v^0) \\
            h &= ReLU(W^AA+b_A) \\
            \theta_v &= A + h
        \end{align}
        where $W^Q \in \mathbb{R}^{d\times d_q}, W^K \in \mathbb{R}^{d\times d_k}, W^V \in \mathbb{R}^{d\times d_v}$ are randomly initialized projection matrices. We set $d_q=d_k=d_v=d/h$. $h$ is the number of heads of attention layer. 
        The translated textual feature $a'$ is computed as follows and $Q_t \in \mathbb{R}^{M \times d}$ is a learnable textual query which shared across samples.
        \begin{equation}
           a' = softmax(\frac{(W^QQ_t)^T(W^K\theta_v)}{\sqrt{d}})(W^V\theta_v) 
        \end{equation}

    \subsection{Training Objective}
    \label{sec:trainingloss}
        \textbf{Sentiment Classification Loss.}
        We compute the loss for sentence classification using cross-entropy loss. $s_i$ represents the predicted sentiment, while $\bar{s_i}$ represents the sentiment label obtained after transformation based on the Algorithm \ref{alg:label}.
        \begin{equation}
            \mathcal{L}_s=CE(\bar{s_n}, s_n) = -\sum_{i=1}^{n} \bar{s_i} log(s_i)
        \end{equation}
        
        \textbf{Semantic Consistency Loss.} 
        The semantic consistency loss serves the purpose of guaranteeing the cross-modal translation of desired features. The consistency loss $\mathcal{L}_c$ for translation is as follows. $D_{KL}$ denotes the Kullback-Leibler Divergence.
        \begin{equation}
             \mathcal{L}_c = D_{KL}(a'||a^0) 
                = -\sum_{i=1}^{n} a'log\frac{a^0}{a'}
        \end{equation}





        \textbf{Overall Objective.}
        To better guide the model during training, our training objective consists of two components. 
        (1) $\mathcal{L}_s$ for the sentiment classification. 
        (2) $\mathcal{L}_c$ in the cross-modal translation-based alignment method is employed to ensure semantic consistency.  
        The learning objective $\mathcal{L}$ is as follows and $\beta$ is the hyperparameter.
            \begin{equation}
                \mathcal{L} = \mathcal{L}_s + \beta \mathcal{L}_c 
                \label{eqt:loss}
            \end{equation}

        
            

\section{Experimental Setting}
    \subsection{Datasets, Metrics and Implement Details}
    \label{sec:detail1}
        Following the previous works \citep{DTCA, DQPSA, AoM, CMMT}, we consider two well-constructed MABSA datasets, mainly consisting of reviews on Twitter, including text and image. These datasets are Twitter2015 and Twitter2017, originally provided by \citet{TwitterPropose} for multimodal named entity recognition and annotated with the sentiment polarity for each aspect by \citet{TwitterLabel}. The statistics of these datasets are shown in Table \ref{tab:dataset}.
        An aspect is regarded as correctly predicted only if the aspect term and polarity respectively match the ground-truth aspect term and corresponding polarity. 
        
        
        For all experiments, the weights of DeBERTa and ViT are respectively initialized from pretrained DeBERTa-base and Vit-base-patch16-224-in21k. The hidden size $d$ is 768, the number of heads in cross-modal self-attention is 8, patch size $P$ is 14.
        Besides, AdamW optimizer \citep{AdamW} with a base learning rate of $2e^{-5}$ and warmup decay of 0.1 is used. The maximum text length and batch size are set to 60 and 16. 
        The Pytorch version is 2.0 and all experiments are conducted on RTXA800. Our experiments require 24GB of VRAM and the time needed to run 150 epochs is approximately 3h.

    \subsection{Baseline}
        To evaluate the proposed model, we select multiple representative advanced methods for comparison: 
        (1) \textit{SPAN} \citep{SPAN} uses span-based identification to locate the positions of targets in the text, effectively addressing the issue of overlapping targets. 
        (2) \textit{D-GCN} \citep{DGCN} , based on directional convolutional networks, jointly tackles aspect extraction and sentiment analysis  while leveraging syntactic information. It effectively utilizes the dependencies among words. 
        (3) \textit{BART} \citep{BART} adapts the MABSA task to BART by redefining it as an index generation problem.
        (4) \textit{UMT-collapse} \citep{UMT} unifies aspect extraction and sentiment analysis into a single process, leveraging a directional graph convolutional network architecture and introducing additional syntactic information to aid in understanding input data. 
        (5) \textit{OSCGA-collapse} \citep{OSCGA}  emphasizes the value of finer-grained information in aiding MABSA tasks, further leveraging object-level image details and character-level text information. 
        (6) \textit{JML} \citep{JML} employs a hierarchical approach to comprehend multimodal information and connects the tasks of MATE and MASC. 
        (7) \textit{VLP} \citep{VLP} is a specialized Vision-Language Pre-training framework designed for MABSA. It features a unified multimodal encoder-decoder architecture for both pretraining and downstream tasks.
        (8) \textit{DTCA} \citep{DTCA} introduces a dual-encoder transformer, and improves cross-modal alignment by introducing auxiliary tasks and an unsupervised approach to enhance text and image representation. 
        (9) \textit{M2DF} \citep{m2df} proposes a Multi-grained Multi-curriculum Denoising Framework (M2DF) based on the concept of curriculum learning to address the negative impact of noisy images. 
        (10) \textit{AoM} \citep{AoM} introduces an aspect-aware attention module to detect aspect-relevant information from text-image pairs, integrating sentiment embedding and graph convolutional networks.
        (11) \textit{DQPSA} \citep{DQPSA} uses a prompt as a dual query module to extract relevant visual information and an energy-based pairwise expert module for improved span prediction.

\begin{table}[tbp]
    \caption{\textbf{MABSA} (Multimodal Aspect-based Sentiment Analysis) results of our model and others with comparison. Bold indicates the best and underline is the second-best performance.}
    \vskip 0.12in
    \small
    \centering
    \begin{tabular}{lccccccrr}
    \toprule
    \multicolumn{1}{c}{\multirow{2}[4]{*}{\textbf{MABSA}}} & \multicolumn{3}{c}{\textbf{Twitter15}} & \multicolumn{3}{c}{\textbf{Twitter17}} & \multicolumn{1}{c}{\multirow{2}[4]{*}{\textbf{Year}}} & \multicolumn{1}{c}{\multirow{2}[4]{*}{\textbf{Publication}}} \\
\cmidrule{2-7}          & \textbf{F1} & \textbf{Precision} & \textbf{Recall} & \textbf{F1} & \textbf{Precision} & \textbf{Recall} &       &  \\
    \midrule
    SPAN~\cite{SPAN}        & 53.8  & 53.7  & 53.9  & 60.6  & 69.6  & 61.7  & 2019 & \multicolumn{1}{l}{ACL} \\
    D-GCN~\cite{DGCN}       & 59.4  & 58.3  & 58.8  & 64.1  & 64.2  & 64.1  & 2020 & \multicolumn{1}{l}{COLING} \\
    BART~\cite{BART}        & 62.9  & 65.0  & 63.9  & 65.2  & 65.6  & 65.4  & 2020 & \multicolumn{1}{l}{ACL} \\
    RoBERTa~\cite{Roberta}  & 63.3  & 62.9  & 63.7  & 65.6  & 65.1  & 66.2  & 2020 & \multicolumn{1}{l}{arXiv} \\
    UMT~\cite{UMT}          & 59.8  & 58.4  & 61.4  & 62.4  & 62.3  & 62.4  & 2020 & \multicolumn{1}{l}{ACL} \\
    OSCGA~\cite{OSCGA}      & 62.5  & 61.7  & 63.4  & 63.7  & 63.4  & 64.0  & 2020 & \multicolumn{1}{l}{ACM MM} \\
    JML~\cite{JML}          & 64.1  & 65.0  & 63.2  & 66.0  & 66.5  & 65.5  & 2021 & \multicolumn{1}{l}{EMNLP} \\
    VLP~\cite{VLP}          & 65.1  & 68.3  & 66.6  & 66.9  & 69.2  & 68.0  & 2022 & \multicolumn{1}{l}{ACL} \\
    CMMT~\cite{CMMT}        & 66.5  & 64.6  & 68.7  & 68.5  & 67.6  & 69.4  & 2022 & \multicolumn{1}{l}{IPM} \\
    M2DF~\cite{m2df}        & 67.6  & 67.0  & 68.3  & 68.3  & 67.9  & 68.8  & 2023 & \multicolumn{1}{l}{EMNLP} \\
    DTCA~\cite{DTCA}        & 68.4  & 67.3  & \underline{69.5}  & 70.4  & 69.6  & \underline{71.2}  & 2022 & \multicolumn{1}{l}{ACL} \\
    AoM~\cite{AoM}          & 68.6  & 67.9  & 69.3  & 69.7  & 68.4  & 71.0  & 2023 & \multicolumn{1}{l}{ACL} \\
    DQPSA~\cite{DQPSA}      & \underline{71.9}  & \underline{71.7}  & \textbf{72.0}  & \underline{70.6}  & \underline{71.1}  & 70.2  & 2024 & \multicolumn{1}{l}{AAAI} \\
    \midrule
    Llama2~\cite{llama2} & 54.3  & 53.6  & 55.0  & 58.2  & 57.6  & 58.8  & 2023  &  \\
    ChatGPT3.5~\cite{ChatGPT} & 51.4  & 50.9  & 51.9  & 55.8  & 55.6  & 56.1  & 2023  &  \\
    \midrule
    \textbf{Ours} & \textbf{74.1 } & \textbf{76.3 } & \textbf{72.0}  & \textbf{73.7 } & \textbf{74.6 } & \textbf{72.8 } &       &  \\
    \bottomrule
    \end{tabular}%
  \label{tab:result}%
\end{table}%

\section{Results and Analysis}        
    \subsection{MABSA Results}
        Table \ref{tab:result} summarizes the comparative results of the proposed PTA model against several previous methods in terms of precision (P), recall (R), and F1-score. As indicated, the proposed model outperforms all the baseline models. Compared with the multimodal baseline with the best performance, i.e. DQPSA, PTA still shows absolute F1-score increases of 2.2\% and 3.1\%. Compared with text-based models, PTA provides far better results. The PTA F1 outperforms RoBERTa by 10.8\% and 8.1\% respectively on Twitter-2015 and Twitter-2017. This indicates that the proposed method could enable PTA to learn an appropriate representation for MABSA. We also provide a case study for further illustrating the superiority of PTA (\S \ref{sec:case}).

    \subsection{MATE and MASC Results}
    \begin{table}[tbp]
  \centering
  \small
  \caption{\textbf{MATE} (Multimodal Aspect Term Extraction) results.}
    \begin{tabular}{lccccccrr}
    \toprule
    \multicolumn{1}{c}{\multirow{2}[4]{*}{\textbf{MATE}}} & \multicolumn{3}{c}{\textbf{Twitter15}} & \multicolumn{3}{c}{\textbf{Twitter17}} & \multicolumn{1}{c}{\multirow{2}[4]{*}{\textbf{Year}}} & \multicolumn{1}{c}{\multirow{2}[4]{*}{\textbf{Publication}}} \\
\cmidrule{2-7}          & \textbf{F1} & \textbf{Precision} & \textbf{Recall} & \textbf{F1} & \textbf{Precision} & \textbf{Recall} &       &  \\
    \midrule
    UMT~\cite{UMT}   & 79.7  & 77.8  & 81.7  & 86.7  & 86.7  & 86.8  & \multicolumn{1}{c}{2020} & \multicolumn{1}{l}{ACL} \\
    OSCGA~\cite{OSCGA} & 81.9  & 81.7  & 82.1  & 90.4  & 90.2  & 90.7  & \multicolumn{1}{c}{2020} & \multicolumn{1}{l}{ACM MM} \\
    JML~\cite{JML} & 82.4  & 83.6  & 81.2  & 91.4  & 92.0  & 90.7  & \multicolumn{1}{c}{2021} & \multicolumn{1}{l}{EMNLP} \\
    VLP~\cite{VLP} & 85.7  & 83.6  & 87.9  & 91.7  & 90.8  & 92.6  & \multicolumn{1}{c}{2022} & \multicolumn{1}{l}{ACL} \\
    CMMT~\cite{CMMT}  & 85.9  & 83.9  & \underline{88.1}  & 93.1  & 92.2  & \underline{93.9}  & \multicolumn{1}{c}{2022} & \multicolumn{1}{l}{IPM} \\
    M2DF~\cite{m2df} & 86.3  & 85.2  & 87.4  & 92.4  & 91.5  & 93.2  & \multicolumn{1}{c}{2023} & \multicolumn{1}{l}{EMNLP} \\
    AoM~\cite{AoM} & 86.2  & 84.6  & 87.9  & 92.3  & 91.8  & 92.8  & \multicolumn{1}{c}{2023} & \multicolumn{1}{l}{ACL} \\
    DQPSA~\cite{DQPSA} & \underline{87.7} & \textbf{88.3 } & 87.1  & \textbf{94.3 } & \textbf{95.1 } & 93.5  & \multicolumn{1}{c}{2024} & \multicolumn{1}{l}{AAAI} \\
    \midrule
    \textbf{Ours} & \textbf{87.8 } & \underline{87.2}  & \textbf{88.3 } & \underline{93.2}  & \underline{92.4}  & \textbf{94.0 } &       &  \\
    \bottomrule
    \end{tabular}%
  \label{tab:mate}%
\end{table}%

    \vspace{-0.3cm}

\begin{table}[tbp]
  \centering
  \small
  \caption{\textbf{MASC} (Multimodal Aspect-oriented Sentiment Classification) results.}
    \begin{tabular}{lcccccr}
    \toprule
    \multicolumn{1}{c}{\multirow{2}[4]{*}{\textbf{MASC}}} & \multicolumn{2}{c}{\textbf{Twitter15}} & \multicolumn{2}{c}{\textbf{Twitter17}} & \multirow{2}[4]{*}{\textbf{Year}} & \multicolumn{1}{c}{\multirow{2}[4]{*}{\textbf{Publication}}} \\
\cmidrule{2-5}          & \textbf{Acc} & \textbf{F1} & \textbf{Acc} & \textbf{F1} &       &  \\
    \midrule
    ESAFN~\cite{ESAFN} & 73.4  & 67.4  & 67.8  & 64.2  & 2020  & \multicolumn{1}{l}{TASLP} \\
    TomBERT~\cite{TOMBERT} & 77.2  & 71.8  & 70.5  & 68.0    & 2019  & \multicolumn{1}{l}{IJCAI} \\
    CapTrBERT~\cite{CapTrBERT} & 78.0    & 73.2  & 72.3  & 70.2  & 2021  & \multicolumn{1}{l}{MM} \\
    JML~\cite{JML} & 78.7  & -     & 72.7  & -     & 2021  & \multicolumn{1}{l}{EMNLP} \\
    VLP-MABSA~\cite{VLP} & 78.6  & 73.8  & 73.8  & 71.8  & 2022  & \multicolumn{1}{l}{ACL} \\
    CMMT~\cite{CMMT} & 77.9  & -     & 73.8  & -     & 2022  & \multicolumn{1}{l}{IPM} \\
    AoM~\cite{AoM} & \underline{80.2}  & \underline{75.9}  & \underline{76.4}  & \underline{75.0}    & 2023  & \multicolumn{1}{l}{ACL} \\
    \midrule
    VisualGLM-6B~\cite{VisualGLM} & -     & 66.8  & -     & 54.5  & 2022  & \multicolumn{1}{l}{ACL} \\
    ChatGPT-3.5~\cite{ChatGPT} & -     & 66.3  & -     & 58.9  & 2023  &  \\
    \midrule
    \textbf{Ours} & \textbf{84.7} & \textbf{78.5} & \textbf{79.8} & \textbf{75.3} &       &  \\
    \bottomrule
    \end{tabular}%
  \label{tab:masc}%
\end{table}%

        In the MABSA task, our recall is lower than our precision on both datasets. This is due to the error propagation issue inherent in the pipeline approach. Errors in the aspect predictions during the MATE stage lead to fewer correct aspects being passed to the MASC stage, resulting in lower recall. 
        This issue is more pronounced on Twitter15 (P:76.3 and R:72.0), where the F1 score for the MATE stage is only 87.8. In contrast, for the Twitter17 dataset (P:74.6 and R:72.8), the MATE F1 is 93.2, leading to more accurate aspect predictions and thus a less severe recall-precision discrepancy.
        
        As shown in Table \ref{tab:mate}, our model performs worse than DQPSA~\citep{DQPSA} on MATE task, but the overall performance on the MABSA task is better, indicating that our approach to utilizing image information is more effective. Given the same ability to recognize aspects, our proposed pipeline method leverages image information better for sentiment classification. Additionally, we attribute the higher mate performance of DQPSA to its pre-training on both datasets, which incurs significant training costs.

    \subsection{Ablation Study}
        We conduct ablation experiments to evaluate the effectiveness of each proposed component, with the results shown in Table \ref{tab:ablation}. The experiments demonstrate that our proposed TBA alignment method (70.5->74.1) and the pipeline framework (70.0->74.1) both significantly improve the performance on the MABSA task. When both methods are combined, the improvement is even more significant (67.5->74.1). It is worth noting that in the ablation experiments testing without the pipeline, we cannot obtain the predicted aspect, so we use the sentence to interact with the image instead.
        Additionally, as shown in Table \ref{tab:encoder}, we conduct a thorough exploration of the encoders used in our model and conclude that DeBERTa better meets the requirements of our task.

\begin{table}[htbp]
  \centering
  \small
  \caption{Ablation of our method. TBA means the translation-based alignment method(\S \ref{sec:TBAM}) and Pipeline means the proposed pipeline framework (\S \ref{pipeline}) for MABSA.}
    \begin{tabular}{lccc}
    \toprule
            & \textbf{F1}    & \textbf{Precision} & \textbf{Recall} \\
    \midrule
    \textbf{Ours}        & 74.1  & 76.3  & 72.0  \\
    w/o TBA      & 70.5  & 70.6  & 70.4  \\
    w/o Pipeline & 70.0  & 70.2  & 69.8  \\
    w/o TBA+Pipeline & 67.5  & 66.0  & 68.9  \\
    \bottomrule
    \end{tabular}%
  \label{tab:ablation}%
\end{table}%



    \subsection{The Experiment Supporting Our Observations and Analyses}
        \label{sec:observation-experiment}
        We experimentally validate and analyze our three observations (\S \ref{sec:intro}) regarding MABSA methods .
        \begin{figure}[htbp]
            \vspace{-10pt}
            \centering
            \subfloat[Shared encoder]{\includegraphics[height=.29\linewidth]{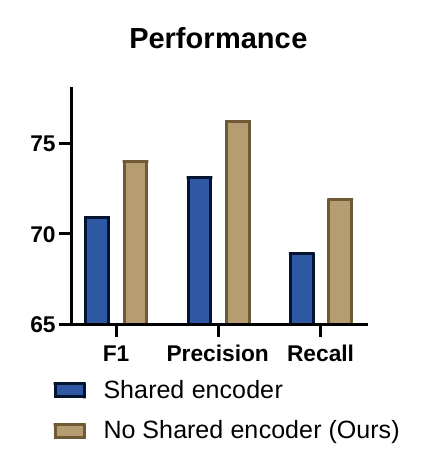}\label{fig:shared}}
            \subfloat[Interaction features]{\includegraphics[height=.29\linewidth]{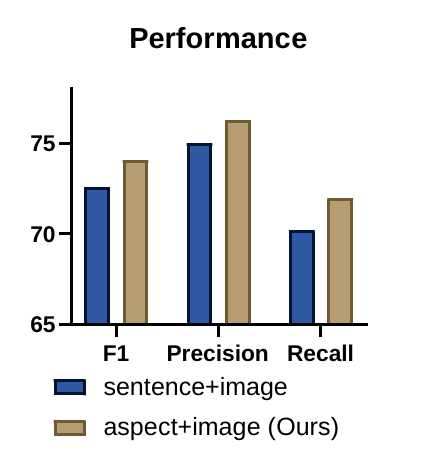}\label{fig:interaction}}
            \subfloat[Image ablation]{\includegraphics[height=.29\linewidth]{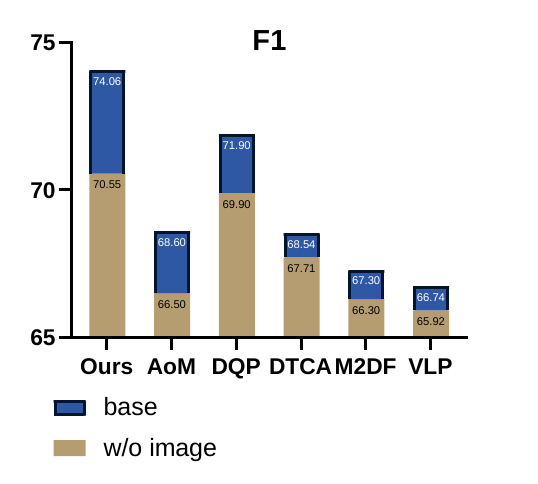}\label{fig:image}}
            \caption{Experimental analysis on Twitter15. 
            (a) The performance of MATE and MASC with shared and \textbf{not} shared encoders.
            (b) The impact of selecting textual features: entire sentence or predicted aspect, to align with image.
            (c) F1 result of different methods with and w/o image.}
        \end{figure}

        \textbf{(a) The MATE and MASC models have different requirements for features.}
        Existing methods often assume that joint models can mutually benefit each other by modeling the interactions between the two tasks (MATE and MASC), thereby accomplishing the MABSA task. In this section, we investigate whether sharing the two representation encoders can improve performance. As shown in Figure \ref{fig:shared}, \textit{simply sharing the encoders hurts MABSA's performance}. We believe this is because the inputs and goals of the two tasks are different, and they \textit{require different features to predict aspects and sentiments}, respectively. 
        Consequently, using a pipeline approach to solve the MABSA task allows the model to tackle the two tasks separately, thereby learning better task-specific features.

        \textbf{(b) The aspect of MATE results is crucial for utilizing images.}
        As shown in Figure \ref{fig:interaction}, in the selection of input features for the alignment(\S \ref{sec:TBAM}), we test using aspects or text to interact with images separately. The results show that \textit{using predicted aspects with images could obtain more effective visual information}. Compared to sentence, which may contain redundant information, aspects could precisely describe key regions or attributes of the image. This highlights the importance of predicted aspects in solving the MABSA and demonstrates the superiority of the pipeline approach.

        \textbf{(c) Images play a trivial role in previous MABSA methods.}
        As shown in Figure \ref{fig:image}, we conduct image ablation experiments on our method and report the performance of different methods with and without image. 
        Previous methods have shortcomings in image utilization:
        After removing images, the performance differences of DTCA~\citep{DTCA}, M2DF~\citep{m2df}, VLP~\citep{VLP} and DQPSA~\citep{DQPSA} are 0.83, 1.00, and 0.82, 2.00, respectively. 
        Even if the aspect is determined, the high noise in images from social media data also demands a more effective approach to image utilization. AoM~\citep{AoM} uses Spacy tagging to extract candidate aspects, but its image impact remains low at 2.10.
        It can be seen the superiority of our image utilization method, with an ablation impact of 3.51 (74.06 -> 70.55).

    \subsection{The influence of hyperparameters}
    \label{sec:detail2}
        \begin{figure}[htbp]
            \centering
            \subfloat[Model parameters(/M)]{\includegraphics[height=.26\linewidth]{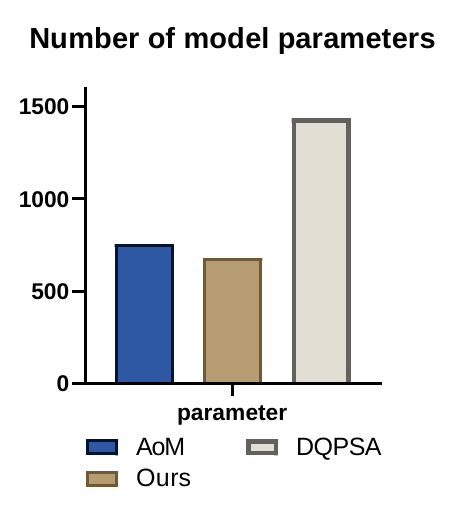}\label{fig:params}}
            \subfloat[Effect of batch size]{\includegraphics[height=.26\linewidth]{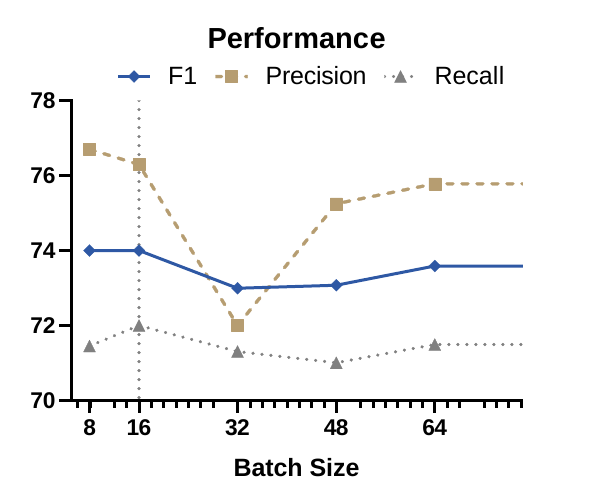}\label{fig:bs}}
            \subfloat[Effect of loss weight $\beta$]{\includegraphics[height=.26\linewidth]{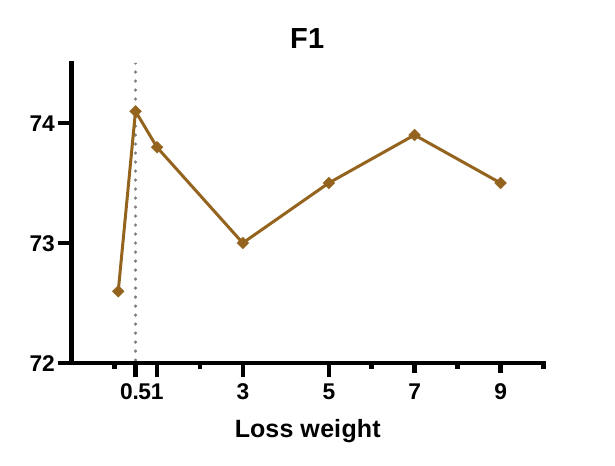}\label{fig:loss}}
            \caption{Experimental analysis of hyperparameters on the Twitter15. }
        \end{figure}
    We conduct detailed analysis experiments in our paper:
    \textbf{(1) Model parameters}.
        We estimate the parameters required for the entire training process of two previous representative methods, AOM~\citep{AoM} and DQPSA~\citep{DQPSA}. Experiments show that our method uses fewer parameters. Specifically, the previous two methods perform pre-training on TRC~\cite{TRC} and COCO2017~\cite{coco} to achieve better performance, resulting in higher training costs. Our method does not require pre-training on any additional datasets. 
        \textbf{(2) Effect of batch size}. We test multiple batch sizes, and the performance results are shown in Figure \ref{fig:bs}. Ultimately, we chose 16 as the batch size.
        \textbf{(3) Effect of loss weight}. It is used to evaluate the weight $\beta$ of the two losses $\mathcal{L}_s$ and $\mathcal{L}_c$  in the section \ref{sec:trainingloss} and we set $\beta = 0.5$.

\section{Conclusion}

    This paper addresses the limitations of joint prediction models in multimodal aspect-based sentiment analysis (MABSA) by proposing a pipeline framework. By first identifying the aspect through MATE and then aligning it with relevant image patches for MASC, our method achieves more precise and effective utilization of image data.
    We highlight three key observations: (a) MATE and MASC have different feature requirements, with MATE focusing on token-level features and MASC on sequence-level features; (b) the aspect identified by MATE is crucial for effective image utilization; and (c) previous MABSA methods struggle with high-noise image data.
    To address these challenges, we propose the pipeline framework that first predicts the aspect and then uses translation-based alignment (TBA) to enhance semantic consistency between text and images.  
    Our extensive experiments on public datasets demonstrate that our method outperforms existing models by a large margin, including large language models like Llama2, ChatGPT, and VisualGLM6B. 
    Overall, we highlight the advantages of a pipeline framework over joint prediction models in MABSA. Our observations and the proposed methods provide insights for future research and applications in this field.

\clearpage
\bibliographystyle{plainnat}
\bibliography{main}
\clearpage

\appendix

\section{Appendix / supplemental material}
    \subsection{Limitation}
    \label{sec:limitation}
        Our method may also face the issue of error accumulation, the errors from the upstream MATE task in the pipeline may propagate to the downstream MASC task. When MATE predicts the aspect incorrectly, the results of MASC are likely to be incorrect. To mitigate the error accumulation problem, we adopt two strategies:

        1. As shown in the algorithm \ref{alg:label}, we employ fuzzy matching between MATE outputs and labels during training to increase the amount of available training data for MASC. When the similarity exceeds a certain threshold, we include them as training data. For instance, "\# Messi" and "Messi" are considered the same aspect, and we label them accordingly based on the dataset annotations.

        2. We use Spacy\footnote{https://github.com/explosion/spaCy} to perform part-of-speech tagging on MATE predictions. Since aspects are typically nouns, we filter out the predictions that are not tagged as nouns.

    \subsection{Case Study}
    \label{sec:case}
    \begin{table*}[htbp]
\centering
\caption{Predictions of different methods on four test samples}
\vskip 0.12in
\setlength{\belowcaptionskip}{-0.3cm}
\setlength{\abovecaptionskip}{0.1cm}
\tiny
    \begin{tabular}{p{0.8cm}p{2.7cm}p{2.7cm}p{2.7cm}p{2.7cm}}
    \toprule
    Image &
    \makebox[2.4cm][c]{\begin{minipage}{0.07\textwidth}
        \hbox{\hspace{-0.1em} \includegraphics[width=15mm, height=13.0mm]{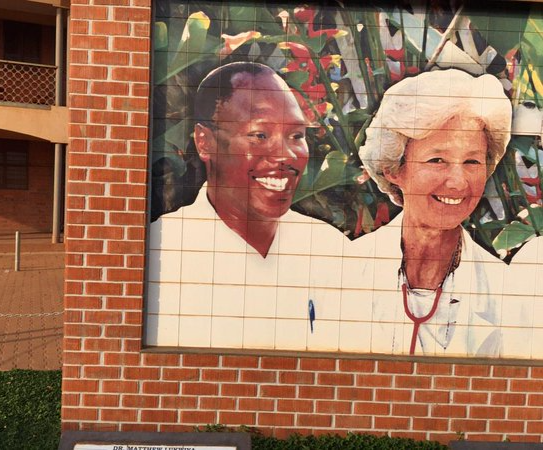}}
      \end{minipage}\vspace{0.3em}}
    &
    \makebox[2.4cm][c]{\begin{minipage}{0.07\textwidth}
        \hbox{\centering \hspace{-0.1em} \includegraphics[width=14mm, height=13.0mm]{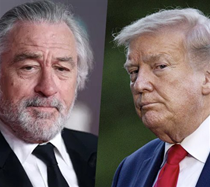}}
      \end{minipage}\vspace{0.3em}}
    &
    \makebox[2.4cm][c]{\begin{minipage}{0.07\textwidth}
        \hbox{\hspace{-0.1em} \includegraphics[width=15mm, height=13.0mm]{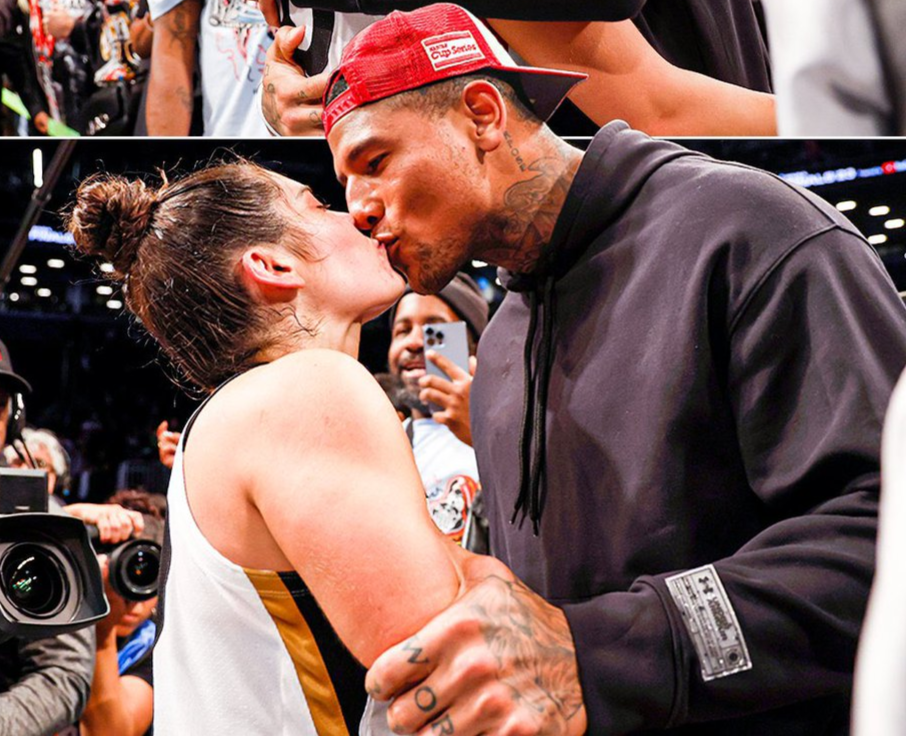}}
      \end{minipage}\vspace{0.3em}}
    &
    \makebox[2.4cm][c]{\begin{minipage}{0.07\textwidth}
        \hbox{\hspace{-0.1em} \includegraphics[width=16mm, height=13.0mm]{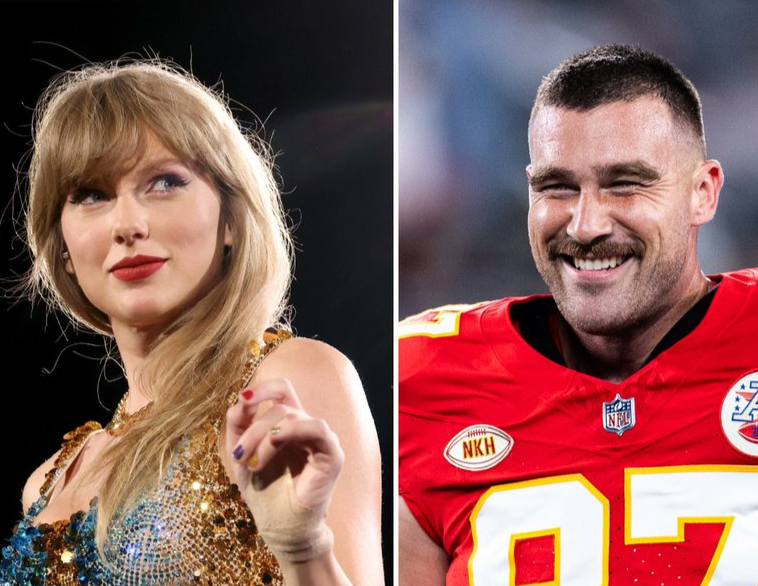}}
      \end{minipage}\vspace{0.3em}}
    \\
    \multirow{4}{*}{Text} &
    (a) What do health heroes look like? Dr Lucille Corti died AIDS 1996, Dr Lukwiya died Ebola 2000 
    &
    (b) In a devastating speech, iconic actor Robert Niro declare that Donald Trump will be failure.
    &
    (c) Kelsey Plum and Devin Booker share a beautiful moment.   
    &
    (d) Breaking news: Renowned singer Taylor will be attending the Travis Kelce Game vs Broncos in KC Thursday! 
    \\
    \midrule
    \multirow{4}{*}{Ground Truth} 
    & (Dr Lucille Corti, \textcolor{teal}{Positive}) & (Robert Niro, \textcolor{violet}{Neutral}) & (Kelsey Plum, \textcolor{teal}{Positive}) & (Taylor, \textcolor{teal}{Positive}) \\
    & (AIDS, \textcolor{purple}{Negative}) & (Donald Trump, \textcolor{purple}{Negative}) & (Devin Booker, \textcolor{teal}{Positive}) & (Travis Kelce, \textcolor{violet}{Neutral}) \\
    & (Dr Lukwiya, \textcolor{teal}{Positive}) & ~ & ~ & (Broncos, \textcolor{violet}{Neutral}) \\
    & (Ebola, \textcolor{purple}{Negative}) & ~ & ~ & \\
    \cmidrule{2-5}
    \multirow{4}{*}{DeBERTa~\cite{DeBERTa}} 
    & \textcolor{red}{$\times$}(Lucille Corti, \textcolor{violet}{Neutral}) & \textcolor{red}{$\times$}(Robert, \textcolor{purple}{Negative}) & \textcolor{blue}{$\checkmark$}(Kelsey Plum, \textcolor{teal}{Positive}) & \textcolor{blue}{$\checkmark$}(Taylor, \textcolor{teal}{Positive}) \\
    & \textcolor{blue}{$\checkmark$}(AIDS, \textcolor{purple}{Negative}) & \textcolor{blue}{$\checkmark$}(Donald Trump, \textcolor{purple}{Negative}) & \textcolor{red}{$\times$} - & \textcolor{red}{$\times$}(Game, \textcolor{violet}{Neutral})  \\
    & \textcolor{red}{$\times$}(Dr Lukwiya, \textcolor{violet}{Neutral}) & ~ & ~ & \textcolor{blue}{$\checkmark$}(Broncos, \textcolor{violet}{Neutral}) \\
    & \textcolor{blue}{$\checkmark$}(Ebola, \textcolor{purple}{Negative}) & ~ & ~ &   \\
    \cmidrule{2-5}
    \multirow{4}{*}{DTCA~\cite{DTCA}} 
    & \textcolor{red}{$\times$}(Corti, \textcolor{teal}{Positive}) & \textcolor{red}{$\times$}(Robert Niro, \textcolor{purple}{Negative}) & \textcolor{red}{$\times$}(Kelsey Plum, \textcolor{violet}{Neutral}) & \textcolor{blue}{$\checkmark$}(Taylor, \textcolor{teal}{Positive}) \\
    & \textcolor{blue}{$\checkmark$}(AIDS, \textcolor{purple}{Negative}) & \textcolor{blue}{$\checkmark$}(Donald Trump, \textcolor{purple}{Negative}) & \textcolor{blue}{$\checkmark$}(Devin Booker, \textcolor{teal}{Positive}) & \textcolor{red}{$\times$}(Travis Kelce, \textcolor{violet}{Neutral}) \\
    & \textcolor{red}{$\times$}(Lukwiya, \textcolor{violet}{Neutral}) & ~ & ~ & \textcolor{red}{$\times$}(Broncos, \textcolor{teal}{Positive}) \\
    & \textcolor{blue}{$\checkmark$}(Ebola, \textcolor{purple}{Negative}) & ~ & ~ &  \\
    \cmidrule{2-5} 
    \multirow{3}{*}{Ours} 
    & \textcolor{blue}{$\checkmark$}(Dr Lucille Corti, \textcolor{teal}{Positive}) & \textcolor{blue}{$\checkmark$}(Robert Niro, \textcolor{violet}{Neutral}) & \textcolor{blue}{$\checkmark$}(Kelsey Plum, \textcolor{teal}{Positive}) & \textcolor{blue}{$\checkmark$}(Taylor, \textcolor{teal}{Positive})  \\
    & \textcolor{blue}{$\checkmark$}(AIDS, \textcolor{purple}{Negative}) & \textcolor{blue}{$\checkmark$}(Donald Trump, \textcolor{purple}{Negative}) & \textcolor{blue}{$\checkmark$}(Devin Booker, \textcolor{teal}{Positive}) & \textcolor{blue}{$\checkmark$}(Travis Kelce, \textcolor{violet}{Neutral}) \\
    & \textcolor{blue}{$\checkmark$}(Dr Lukwiya, \textcolor{teal}{Positive}) & ~ & ~ & \textcolor{blue}{$\checkmark$}(Broncos, \textcolor{violet}{Neutral}) \\
    & \textcolor{blue}{$\checkmark$}(Ebola, \textcolor{purple}{Negative}) & ~ & ~ & \\
    \bottomrule
    \end{tabular}
    \label{tab:case_study}
\end{table*}

    To better understand the advantage of our method, we select the representative samples with predictions of DeBERTa and DTCA in Table \ref{tab:case_study}. First, we can see that DeBERTa and DTCA heavily rely on text to recognize aspects that may introduce harmful textual prior. 
    For example, in Case (a) and (d), both DeBERTa and DTCA get inaccurate prediction on "Dr Lukwiya" and "Travis Kelce". DeBERTa makes an incorrect sentiment judgment due to the absence of image assistance. Meanwhile, DTCA makes errors in both aspect and sentiment judgment, indicating a deficiency in its comprehension of multimodal information. 
    In contrast, we leverage a translation method to guarantee modality alignment. We treat both modalities equitably in translation, emphasizing the content of each modality as input and output. Our method excels in achieving enhanced coordination between visual and textual information, harnessing the potential of visual data as complementary information for textual content. Thus, our method could recover their true sentiment of "Dr Lukwiya" and "Travis Kelce". 
    
    Furthermore, the understanding of multimodal context is superficial in DTCA. 
    In cases (b) and (c), DTCA correctly predicts the aspect, including "Robert Niro" and "Kelsey Plum", but makes errors in sentiment judgment. This may be attributed to the inadequate alignment between images and text. This indicates that relying solely on text for sentiment assessment can lead to misjudgments. Within our framework, we adopt a translation-based approach to comprehend text-image pairs. This approach places a stronger emphasis on content comprehension with the guidance of translation, equipping our model with robust reasoning capabilities.

\begin{table}[htbp]
  \centering
  \small
  \caption{Performance on the Twitter15 after using different encoders as our textual encoder. We take DeBERTa as our encoder.}
    \begin{tabular}{lccc}
    \toprule
          & \textbf{F1} & \textbf{Precision} & \textbf{Recall} \\
    \midrule
    BERT~\cite{BERT}  & 68.2  & 70.4  & 66.2  \\
    T5~\cite{T5}    & 71.5  & 71.6  & 71.6  \\
    RoBERTa~\cite{Roberta} & 72.6  & 73.8  & 71.3  \\
    BART~\cite{BART}  & 72.7  & 75.2  & 69.8  \\
    DeBERTa~\cite{DeBERTa}\textbf{(Ours)} & \textbf{74.1}  & \textbf{76.3}  & \textbf{72.0}  \\
    \bottomrule
    \end{tabular}%
  \label{tab:encoder}%
\end{table}%
      
\begin{table}[htbp]
  \centering
  \caption{Statistics of two publicly available datasets collected and annotated from social media platforms (\#S, \#A, \#Pos, \#Neu, \#Neg, MA, MS, Mean and Max denote numbers of sentences, aspects, positive aspects, neural aspects, positive aspects, multi aspects in each sentence, multi sentiments in each sentence, mean length and max length) \cite{DTCA}.}
    \begin{tabular}{ccccccccccc}
    \toprule
    \multicolumn{2}{c}{\textbf{Datasets}} & \textbf{\#S} & \textbf{\#A} & \textbf{\#Pos} & \textbf{\#Neu} & \textbf{\#Neg} & \textbf{MA} & \textbf{MS} & \textbf{Mean} & \textbf{Max} \\
    \midrule
    \multirow{3}[1]{*}{\textbf{Twitter-15}} & Train & 2100  & 3179  & 928   & 1883  & 368   & 800   & 278   & 15    & 35 \\
          & Dev   & 737   & 1122  & 303   & 670   & 149   & 286   & 119   & 16    & 40 \\
          & Test  & 674   & 1037  & 317   & 607   & 113   & 258   & 104   & 16    & 37 \\
    \multirow{3}[1]{*}{\textbf{Twitter-17}} & Train & 1745  & 3562  & 1508  & 1638  & 416   & 1159  & 733   & 15    & 39 \\
          & Dev   & 577   & 1176  & 515   & 517   & 144   & 375   & 242   & 16    & 31 \\
          & Test  & 587   & 1234  & 493   & 573   & 168   & 399   & 263   & 15    & 38 \\
    \bottomrule
    \end{tabular}%
  \label{tab:dataset}%
\end{table}%

    \subsection{Broader Impacts}
    \label{sec:impact}
    Our research highlights three key observations, providing valuable insights for future researchers in the field of sentiment analysis. Additionally, our method improves the accuracy of existing sentiment analysis tasks, facilitating more precise completion of MABSA tasks and further advancing AI development. 
    It is important to note that our model may still exhibit inaccuracies and may misinterpret human emotions.


\subsection{Algorithms}
    \begin{algorithm}[h]
  \caption{Convert MATE predict logits to textual aspect} 
  \begin{algorithmic}[1]
\Require
           $L$: predict label;  $W$: words ids;
\Ensure    $R$: aspect span result;

\For{$i = 1$; $i<len(L)$; $i++$ }
    \State State variables $\theta$ (True or False);
    \State The starting index $a$ and ending index $b$ of the aspect.
    
    \If {$W_{i,j}$ is None and $\theta$} 
        \State add (a,b) into $R_i$
    \EndIf
    \If{$W_{i,j}$ != $W_{i,j-1}$}
        \If{$L_{i,j} == 1$}
            \If{$\theta$}
                \State add (a,b) into $R_i$
            \EndIf
            \State $a$, $b$ = $W_{i,j}$, $W_{i,j}$
            \State $\theta=True$
        \EndIf
        \If{$L_{i,j}==2$ and $\theta$}
            \State $b$ = $W_{i,j}$
        \EndIf
        \If{$L_{i,j}==0$}
            \If{$\theta$}
                \State add (a,b) into $R_i$
            \EndIf
            \State $\theta=False$
        \EndIf
    \EndIf
\EndFor
  
\end{algorithmic}
\label{alg:logits}
\end{algorithm}

    \begin{algorithm}[h]
  \caption{Attach sentiment label based on predicted aspect (MATE)} 
  \begin{algorithmic}[1]
\Require
    \State $k$: Batch Size;
    \State $D=[d_1, d_i, ..., d_k]$: sentence;
    \State $P=[p_1, p_i, ..., p_k]$: predict aspect, where $p_i= [p'_1, p'_2, p'_m]$;
    \State $T=[t_1, t_i, ..., t_k]$: true pairs, where $t_i= [t'_1, t'_2, t'_n]$ and $t'_i = (a,s)$. $a$ is aspect and $s$ is corresponding sentiment;
\Ensure 
Label $l=(a, d_i, \bar{s})$;

\For{$i = 1$; $i<len(P)$; $i++$}
    \State current sentence $d_i$, current predict aspect $p_i$ and current true pairs $t_i$;
    \State \# don't predict any aspect, thus judge the sentiment of whole sentence.
    \If {$len(p_i)==0$} 
        \State \# aspect $a$ is set to be None, $s$ is set to be the sentiment of the first true pair.
        \State add (None, $d_i$, $t'_0.s$) into $l$ 
    \Else 
        \For{$p'_j$ in $p_i$}
            \State Set max similarity $m_s=0$, max value $m_v=None$.  
            \For{$t'_j$ in $t_i$}
                \State Similarity $\gamma$ = sim($p'_j$, $t'_j$ ) 
                \textcolor{blue}{\Comment{sim() is the function to calculate the similarity}}
                \If{$\gamma > m_s$}
                    \State $m_s = \gamma$
                    \State $m_v = t'_j.s$
                \EndIf
            \EndFor
            
            \If{$m_s$ > $\theta$} \textcolor{blue}{\Comment{$\theta$ is the threshold of similarity}}
                \State add ($p'_j$, $d_i$, $t'_j.s$) into $l$
            \Else
                \State add ($p'_j$, $d_i$, $t'_0.s$) into $l$
            \EndIf
        \EndFor
    \EndIf
\EndFor
  
  \end{algorithmic}
  \label{alg:label}
\end{algorithm}

\end{document}